\documentclass[conference]{IEEEtran}
\IEEEoverridecommandlockouts
\usepackage{cite}
\usepackage{amsmath,amssymb,amsfonts}
\usepackage{algorithmic}
\usepackage[utf8]{inputenc}
\usepackage[ruled,longend]{algorithm2e}
\usepackage{textgreek}
\usepackage{amssymb}
\usepackage{graphicx}
\usepackage{textcomp}
\usepackage{url}
\usepackage{xcolor}
\def\BibTeX{{\rm B\kern-.05em{\sc i\kern-.025em b}\kern-.08em
  T\kern-.1667em\lower.7ex\hbox{E}\kern-.125emX}}

\begin{document}

\title{Creating Hierarchical Dispositions of Needs in an Agent\\
\par\rule{\textwidth}{0.5pt}  

\author{\IEEEauthorblockN{1\textsuperscript{st} Tofara Moyo}

Bulawayo , Zimbabwe \\
tofaramoyo@gmail.com, Mazusa AI}

}

\maketitle

\begin{abstract}
We present a novel method for learning hierarchical abstractions that prioritize competing objectives, leading to improved global expected rewards. Our approach employs a secondary rewarding agent with multiple scalar outputs, each associated with a distinct level of abstraction. The traditional agent then learns to maximize these outputs in a hierarchical manner, conditioning each level on the maximization of the preceding level. We derive an equation that orders these scalar values and the global reward by priority, inducing a hierarchy of needs that informs goal formation. Experimental results on the Pendulum v1 environment demonstrate superior performance compared to a baseline implementation.We achieved state of the art results.
\end{abstract}

\bigskip

\section{Introduction}

\bigskip

The complexity of a policy learned by reinforcement learning (RL) algorithms is inherently bounded by the complexity of the reward function. Consequently, significant efforts have been devoted to crafting intricate reward functions that can guide RL agents towards sophisticated behaviors.In contrast, humans and other animals appear to develop complex behaviors through a hierarchical process, wherein an initially simple reward function focused on fundamental drives such as pain avoidance and pleasure seeking serves as the foundation for a layered structure of dispositions. 

\bigskip

Each level in this hierarchy is oriented towards satisfying the preceding levels, ultimately referencing the base reward function.

\bigskip

The mechanisms underlying this process remain unclear. However, if we could induce artificial agents to learn hierarchical reward functions, it would enable the specification of simple base reward functions, allowing the algorithm to autonomously develop complex goals. A hierarchical reward function would confer upon the agent the capacity to pursue intricate objectives.
The hierarchical structure of human needs has been extensively studied, yielding frameworks such as Maslow's Hierarchy of Needs. This hierarchy progresses from fundamental, essential needs to more abstract and complex requirements, including self-actualization.

\bigskip

This paper presents a novel approach to inducing hierarchical reward structures in artificial agents. Our method involves introducing a secondary rewarding agent that parallels the traditional agent, receiving identical state inputs. The rewarding agent features a continuous action output layer, wherein the outputs serve as signals rather than control inputs.

\bigskip

We propose an equation that integrates these signals, yielding a reward signal that is used to reinforce the traditional agent. This framework is designed to elicit a hierarchical organization of needs within the traditional agent, promoting more effective and efficient learning.

\bigskip

Our approach offers two primary benefits: enhanced stability throughout the training process and improved accuracy in the learned policy.

\bigskip

\section{Background}

\subsection{Markov decision process}

We formulate our model of continuous control reinforcement within the framework of a finite Markov Decision Process (MDP). An MDP is defined by the tuple:
$M=\langle S,A,s_{0},r \rangle $ where $S$ denotes the state space $A$ denotes the action space $s_{0} \in S $ denotes the initial state $r(s, a) : S \times A\longrightarrow R$ denotes the reward function, which assigns a scalar value to each state-action pair.At each time step t, the agent selects an action $a_{t+1}$ according to a policy$ \pi:S\longrightarrow A$, which can be either stochastic or deterministic.A stochastic policy is defined as a probability distribution over actions given a state $\pi(a\mid s):S\longrightarrow P(A)$ where $P(A)$ denotes the set of probability distributions over $A$.The objective of the agent is to maximize its future expected reward: $\max \pi\mathbf{E}[\sum_{t=0}^{\infty} r(s_{t},a_{t})]$.

\bigskip
\subsection{Policy Gradient Methods}
Policy gradient methods are a type of reinforcement learning algorithm that learns to optimize the policy directly, rather than learning the value function.
The policy gradient theorem provides the foundation for policy gradient methods. It states that the gradient of the expected cumulative reward with respect to the policy parameters can be computed as:

\bigskip

\begin{equation}
	\nabla_\theta J(\theta) = \mathbb{E}_{\substack{\ s \sim \mu_\pi \\a \sim \pi}} \Big[\nabla_\theta{\log\pi(a|s)}Q_\pi(s,a)\Big],
	\label{eqn:like-grad}
\end{equation}
\bigskip

where $J(\pi_{\theta})$ is the expected cumulative reward
$\pi_{\theta}$ is the policy parameterized by $\theta \tau$ is a trajectory sampled from the policy $s_{t}$ and $a_{t}$ are the state and action at time $t$ $Q\pi\theta(s_{t},a_{t})$ is the action-value function
$\bigtriangledown\theta\log\pi\theta(a_{t}\mid s_{t})$ is the gradient of the log-probability of the action.

\subsubsection{Actor-Critic Methods}

Actor-critic methods are a type of policy gradient method that uses an actor to represent the policy and a critic to estimate the value function. The actor and critic are updated simultaneously using the policy gradient theorem. In this work we used a PPO implementation based of an actor critic network.

\subsubsection{Proximal Policy Optimization}

 We implement a policy gradient method using a truncated version of the generalized advantage estimator (GAE). The GAE is computed as:

 \begin{align}
\hat{A}_{t} = \delta_{t} + (\gamma\lambda)\delta_{t+1} + ... + ... + (\gamma\lambda)^{T-t+1}\delta_{T-1}
\end{align}

\bigskip
 
 where
 
 \bigskip
 
\begin{align}
\delta_{t} = r_{t} + \gamma V(s_{t+1}) - V(s_{t}),
\end{align}

\bigskip

The policy is run for $T$ timesteps, with $T$ less than the episode size. We use the standard notation for the discount factor $\gamma$ and GAE parameter $\lambda$.
To perform a policy update, each of $N$ parallel actors collects $T$ timesteps of data. We then construct the surrogate loss on these $NT$ timesteps of data and optimize it using the ADAM algorithm with a learning rate $\alpha$. We use mini-batches of size $m\leq NT$ for $K$ epochs.

\bigskip

We use a combined loss function that includes the policy surrogate, value function error term, and entropy term:

\bigskip

$
    \textit{L}_{t}^{CLIP+VF+S}\\(\theta)=\hat{\mathbb{E}}_t \left[\\
    \textit{L}_{t}^{CLIP}(\theta) -\\
    c_{1} \textit{L}^{VF}_{t}(\theta)+\\
    c_{2} \textit{S}[\pi_{\theta}](s_{t})\\ 
    \right ],
$

\bigskip

where S denotes the entropy bonus, $L_{t}VF$ is the value function squared-error loss, and $c_{1}$ and $c_{2}$ are coefficients for the value function loss and entropy term, respectively.

\section{Hierarchical Reward Functions}

When creating a reward function that fosters a hierarchical structure of dispositions, it is crucial to establish a sensitive relationship between variables. This can be achieved through a sequential approach. The reward function's equation provides insight into this relationship:

\bigskip

$r = R r_{1} + R$

\bigskip

Where $r$ represents the final scalar reward value derived from the sole output of the rewarding agent.
$R$ denotes the global reward at time step $t$.
$r_{1}$ symbolizes the reward component derived from the reward critics sole output.

\bigskip

A closer examination of the equation reveals that the final reward value $r$ can only be increased by considering the impact of $r_{1}$ on $R$. Since $r_{1}$ and $R$ are correlated, actions that optimize $r_{1}$ at the expense of $R$ will lead to suboptimal values of the final reward $r$.

\bigskip

This correlation between $r_{1}$ and $R$ sets up a two-stage hierarchical structure:

\bigskip

Optimizing $r_{1}$: The agent must perform actions that optimize the reward component $r_{1}$.

\bigskip

Optimizing $R$: Simultaneously, the agent must ensure that these actions also optimize the global reward $R$, ultimately leading to an optimal final reward $r$.

\bigskip

This hierarchical structure encourages the agent to develop a nuanced understanding of the relationships between variables and to make decisions that balance competing objectives.

\bigskip

All the while training the critic soley with the global reward R.

\section{Experiments}

\bigskip

This section presents the results of our experimental evaluation of the proposed hierarchical reward function on a continuous control problem from the OpenAI Gym suite: the Pendulum-v1 environment with a low-dimensional state space.

\bigskip

The architecture of our experimental setup for the Pendulum-v1 environment consisted of a neural network with a final layer outputting a 1-dimensional real-valued vector. Our implementation of the Proximal Policy Optimization (PPO) algorithm was based on a publicly available GitHub repository.

\bigskip

For each environment, we trained five models using different random seeds for a fixed total number of time steps. Following completion of training, each model was evaluated over 100 consecutive episodes to assess its performance.

\bigskip

The performance of each model was evaluated using the cumulative reward obtained over the 100 evaluation episodes. This metric provides a comprehensive assessment of the model's ability to maximize the reward function while adapting to the environment's dynamics.

\bigskip

\subsection{Pendulum-v1}

The Pendulum-v0 environment is a well-established continuous control task from the OpenAI Gym suite. The primary objective of this task is to stabilize a pendulum by applying a torque, effectively balancing the pole in an upright position.

\bigskip

The Pendulum-v0 environment is characterized by:
An unbounded, 3-dimensional observation space
A 1-dimensional action space, where actions represent the torque applied to the pendulum
Bounded actions within the interval $[-2,2]$.

\includegraphics[width=0.36\textwidth]{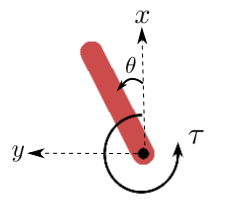}

The agent follows an actor-critic framework. 
The actor $\pi_\theta(a|s)$ consists of a neural network made of 3 fully-connected layers of 64 units each, with $\tanh$ activation functions. 
The output layer has 1 linear neuron.
The critic $V_{\theta_v}(s)$ does not share layers with the actor, but has an equivalent architecture of 3 hidden layers, and one output neuron representing the value function.

\subsection{Reward agent}

The reward agent follows an actor-critic framework. 
The actor $\pi_\theta(a|s)$ consists of a neural network made of 5 fully-connected layers of 64 units each, with $\tanh$ activation functions. The output layer has 3 linear neuron. This is because we wanted to set up a 3 level heirarchy.The equation we used was of the form of the equation we presented earlier but evolved to include a hierarchy of 3 steps. It took the following form.

\bigskip 

$r=R(r_{1}(r_{2}r_{3}+r_{2})+r_{1})+R $

\bigskip

As you can see the equation is a rewrite of the earlier equation, with a replacement of terms.

\bigskip

\section{Results and Discussion}

\bigskip

\subsection{Pendulum-v1}

\bigskip

For the Pendulum-v1 environment, we observe that our method learnt faster, with greater stability and higher rewards than the PPO method without our adjustments.See FIG 2. 

\bigskip

\includegraphics[scale=0.4]{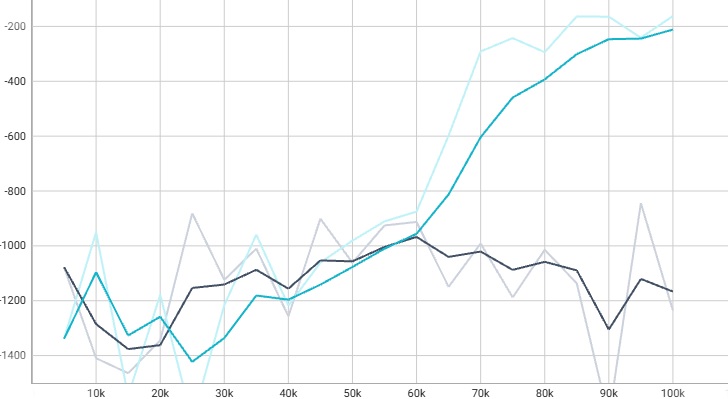}

\bigskip

In blue is our algorithm  vs black, the established method -Average rewards over a 10-episode window for the Pendulum task.

\bigskip

Additionally ,in another run, we were able to beat the state of the art with the TLA model after adapting the github code to include our reward function.The previous state of the art was held by the vanilla version of the TLA algorithm.The results obtained by it were -154 reward points while ours achieved a higher score of -125.Below is a comparison of the two methods results.

\bigskip

\includegraphics{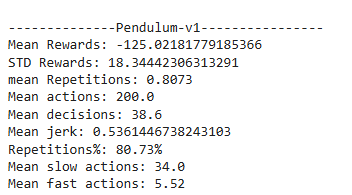}

\bigskip

\includegraphics[scale=0.8]{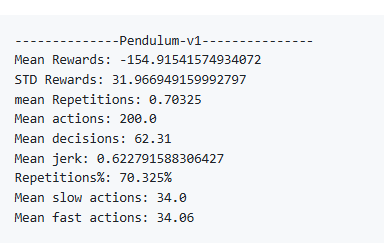}

\bigskip

\section{Future Work}

In our initial approach, we assumed a linear reward function, where scalar values are prioritized and each level is multiplied by the level above, with an additional term. However, this simplistic model may not accurately capture the complexities of real-world systems.

\bigskip

A more comprehensive approach would involve using a graph to model the reward dynamics of the system. In this framework nodes at the same depth would be summed, rather than multiplied, to capture the cumulative effects of different factors and nodes above would be multiplied to represent the hierarchical relationships between different components.This graph-based approach would enable a more nuanced and accurate representation of the reward function.

\bigskip

To implement this graph-based reward modeling, we propose a reward critic architecture that takes the state of the agent as input and outputs a graph representing the reward dynamics. We would then trace each leaf node up to the root, collecting values in an array to form a line-based partial reward function for the traditional agent and sum the rewards over all leaves to obtain the final reward.

\bigskip

To further enhance the reward critic, we can modify it to take into account the actions of the traditional agent, in addition to the state. This would enable the reward critic to evaluate both states and actions, providing a more comprehensive assessment of the agent's behavior.By exploring these graph-based reward modeling and reward critic architectures, we can develop more sophisticated and accurate reward functions that capture the complexities of real-world systems.

\bigskip

\section{Conclusions}

In this study, we conducted a comprehensive evaluation of the effectiveness of hierarchical reward functions in reinforcement learning. Our results demonstrate that agents trained with hierarchical reward functions exhibit faster convergence, improved stability, and higher final rewards compared to agents implementing standard Proximal Policy Optimization (PPO) algorithms.

\bigskip

A comparative analysis of the performance of agents trained with hierarchical reward functions and standard PPO algorithms reveals significant advantages of the former approach. Specifically: Agents trained with hierarchical reward functions exhibit faster convergence rates, achieving optimal performance in fewer iterations
.The stability of agents trained with hierarchical reward functions is improved, with reduced variance in performance across different trials and the final rewards obtained by agents trained with hierarchical reward functions are consistently higher than those achieved by agents implementing standard PPO algorithms.
  
\bigskip

Our results suggest that the proposed method of implementing hierarchical reward functions is effective for simple cases. To further establish the scalability and generalization of this approach, we plan to extend our experiments to more complex environments with intricate dynamics.

\bigskip

We propose to develop more complex graph-based hierarchical reward functions to capture nuanced relationships between different components. This will enable the creation of more sophisticated reward functions that can effectively guide the learning process in complex environments.

\bigskip

A key advantage of the proposed hierarchical reward function approach is the potential for component reuse and transfer learning. By fostering the reuse of components learned early on in the development process, we can accelerate the learning process and improve the overall performance of the agent.

\bigskip

The proposed hierarchical reward function approach has significant implications for complex goal formation and navigation in real-world environments. By emulating the hierarchy of needs exhibited by humans, we can create agents that are capable of navigating complex environments and achieving sophisticated goals.

\bigskip

Future work will focus on extending the proposed hierarchical reward function approach to more complex environments and developing more sophisticated graph-based reward functions.

\vspace{12pt}
\color{red}

\end{document}